\documentclass[letterpaper, 10 pt, conference]{ieeeconf}
\IEEEoverridecommandlockouts
\usepackage[vlined, ruled, boxed, linesnumbered]{algorithm2e}

\SetCommentSty{mycommfont}

\SetKwProg{Function}{}{}{}

\usepackage{graphicx}
\usepackage{graphics}
\usepackage{times}
\usepackage{amsmath}
\usepackage{amssymb}
\usepackage{float}
\usepackage{url}
\usepackage{subfigure}
\usepackage{caption}
\usepackage{multirow}
\usepackage{gensymb}
\usepackage{epstopdf}
\usepackage{wrapfig}
\usepackage{setspace} 
\usepackage[noend]{algpseudocode}
\usepackage{color}
\usepackage{cite}
\usepackage{booktabs}
\usepackage{stackengine}
\usepackage[font={small}]{caption}

\usepackage{tikz}
\usepackage{textcomp}
\usepackage{lipsum}

\newcommand{\bx}{\mathbf{x}}

\newcommand{\bX}{\mathbf{X}}
\newcommand{\bb}{\mathbf{b}}

\newcommand{\bv}{\mathbf{v}}
\newcommand{\bp}{\mathbf{p}}
\newcommand{\bR}{\mathbf{R}}

\newcommand{\bT}{\mathbf{T}}

\title{\LARGE \bf
LVI-SAM: Tightly-coupled Lidar-Visual-Inertial Odometry\\ via Smoothing and Mapping
}

\author{Tixiao Shan, Brendan Englot, Carlo Ratti, and Daniela Rus
\thanks{
\scriptsize{
T. Shan and C. Ratti are with the Department of Urban Studies and Planning, Massachusetts Institute of Technology, USA, {\tt\scriptsize \{shant, ratti\}@mit.edu}. \newline
\indent\; B. Englot is with the Department of Mechanical Engineering, Stevens Institute of Technology, USA, {\tt\scriptsize benglot@stevens.edu}. \newline
\indent\; T. Shan and D. Rus are with the Computer Science \& Artificial Intelligence Laboratory, Massachusetts Institute of Technology, USA, {\tt\scriptsize \{shant, rus\}@mit.edu}.} }
}

\begin{document}

\maketitle
\thispagestyle{empty}
\pagestyle{empty}


\begin{abstract}
We propose a framework for tightly-coupled lidar-visual-inertial odometry via smoothing and mapping, LVI-SAM, that achieves real-time state estimation and map-building with high accuracy and robustness. LVI-SAM is built atop a factor graph and is composed of two sub-systems: a visual-inertial system (VIS) and a lidar-inertial system (LIS). The two sub-systems are designed in a tightly-coupled manner, in which the VIS leverages LIS estimation to facilitate initialization. The accuracy of the VIS is improved by extracting depth information for visual features using lidar measurements. In turn, the LIS utilizes VIS estimation for initial guesses to support scan-matching. Loop closures are first identified by the VIS and further refined by the LIS. LVI-SAM can also function when one of the two sub-systems fails, which increases its robustness in both texture-less and feature-less environments. LVI-SAM is extensively evaluated on datasets gathered from several platforms over a variety of scales and environments. Our implementation is available at \url{https://git.io/lvi-sam}.
\end{abstract}

\section{Introduction}

Simultaneous localization and mapping (SLAM) is a foundational capability required for many mobile robot navigation tasks.
During the last two decades, great success has been achieved using SLAM for real-time state estimation and mapping in challenging settings with a single perceptual sensor, such as a lidar or camera. Lidar-based methods can capture the fine details of the environment at long range. However, such methods typically fail when operating in structure-less environments, such as a long corridor or a flat open field. Though vision-based methods are especially suitable for place recognition and perform well in texture-rich environments, their performance is sensitive to illumination change, rapid motion, and initialization. Therefore, lidar-based and vision-based methods are each often coupled with an inertial measurement unit (IMU) to increase their respective robustness and accuracy. A lidar-inertial system can help correct point cloud distortion and account for a lack of features over short periods of time. Metric scale and attitudes can be recovered by IMU measurements to assist a visual-inertial system. To further improve system performance, the fusion of lidar, camera, and IMU measurements is attracting increased attention.

Our work is most closely related to visual-inertial odometry (VIO), lidar-inertial odometry (LIO), and lidar-visual-inertial odometry (LVIO).
We note that we do not consider non-inertial systems in this paper, though we are aware that there are successful non-inertial lidar-visual systems, such as \cite{graeter2018limo}, \cite{shin2020dvl}. Visual-inertial odometry (VIO) can be grouped into two main categories: filter-based methods and optimization-based methods. Filter-based methods typically use an extended Kalman Filter (EKF) to propagate system states using measurements from a camera and IMU. Optimization-based methods maintain a sliding window estimator and minimize the visual reprojection errors along with IMU measurement error. In our work, we limit our consideration to monocular cameras. 
Among the most popular publicly-available VIO pipelines, MSCKF \cite{mourikis2007multi}, ROVIO \cite{bloesch2015robust} and Open-VINS \cite{geneva2019openvins} are filter-based, and OKVIS \cite{leutenegger2015keyframe}, Kimera \cite{rosinol2020kimera} and VINS-Mono \cite{qin2018vins} are optimization-based. Though OKVIS shows superior performance using a stereo camera, it is not optimized for a monocular camera. VINS-Mono performs non-linear optimization in a sliding window setting and achieves state-of-the-art accuracy with a monocular camera \cite{delmerico2018benchmark}.

Based on their design scheme, lidar-inertial odometry can also be grouped into two main categories: loosely-coupled methods and tightly-coupled methods. LOAM \cite{zhang2017low} and LeGO-LOAM \cite{shan2018lego} are loosely-coupled systems, as the IMU measurements are not used in the optimization step. Tightly-coupled systems, which usually offer improved accuracy and robustness, are presently a major focus of ongoing research \cite{chen2018review}. Among the publicly-available tightly-coupled systems, LIO-mapping \cite{ye2019tightly} adapts the optimization pipeline of \cite{qin2018vins} and minimizes the residuals of IMU and lidar measurements. Because LIO-mapping is designed to optimize all measurements, real-time performance is not achieved. LIO-SAM \cite{shan2020lio}, which bounds computational complexity by introducing a sliding window of lidar keyframes, utilizes a factor graph for joint IMU and lidar constraint optimization. Specifically designed for ground vehicles, LINS \cite{qin2020lins} uses an error-state Kalman filter to correct the state of the robot recursively.

\begin{figure*}[ht]
	\centering
	\includegraphics[width=.99\textwidth]{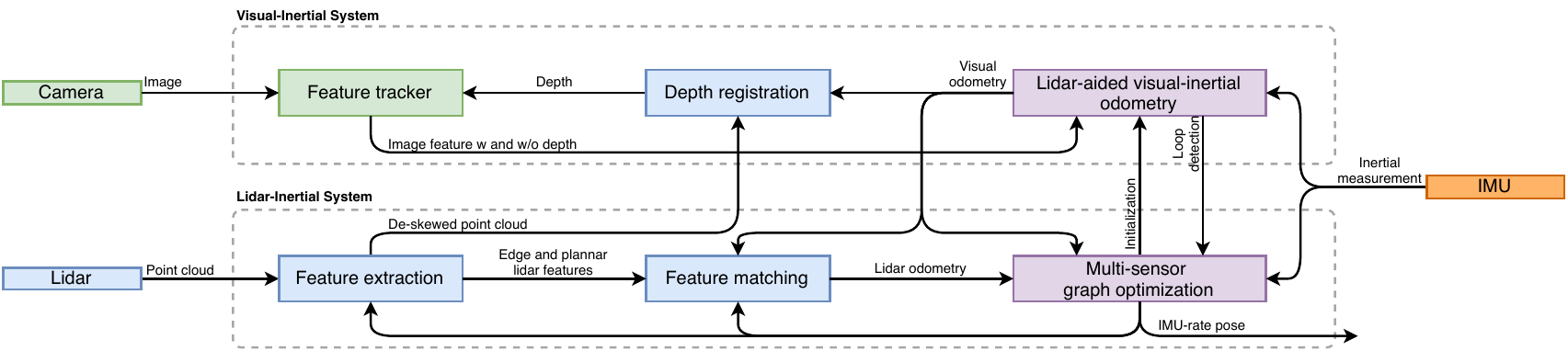}
	\caption{The system structure of LVI-SAM. The system, which receives input from a 3D lidar, a camera and an IMU, can be divided into two sub-systems: a visual-inertial system (VIS) and a lidar-inertial system (LIS). The VIS and LIS can function independently while using information from each other to increase system accuracy and robustness. The system outputs pose estimates at the IMU rate.}
	\label{fig::system-overview}
	\vspace{-5mm}
\end{figure*}

Recently, lidar-visual-inertial systems have attracted increased attention due to their robustness in sensor degraded tasks \cite{debeunne2020review}. \cite{zhang2018laser} proposes a tightly-coupled LVIO system that features a sequential processing pipeline, which solves the state estimation problem from coarse to fine. The coarse estimation starts with IMU prediction and is then further refined by VIO and LIO. \cite{zhang2018laser} currently achieves the state-of-the-art accuracy on the KITTI benchmark \cite{geiger2013vision}. Based on the framework of MSCKF, \cite{zuo2019lic} features online spatial and temporal multi-sensor calibration. The implementations of \cite{zhang2018laser} and \cite{zuo2019lic} are not publicly available. Our work is different from the aforementioned work, as we leverage a factor graph for global optimization, which can regularly eliminate a robot's incurred drift via loop closure detection. 

In this paper, we present a framework for tightly-coupled lidar-visual-inertial odometry via smoothing and mapping, LVI-SAM, for real-time state estimation and mapping. Built atop a factor graph, LVI-SAM is composed of two sub-systems, a visual-inertial system (VIS) and a lidar-inertial system (LIS). The two sub-systems can function independently when failure is detected in one of them, or jointly when enough features are detected.
The VIS performs visual feature tracking and optionally extracts feature depth using lidar frames. The visual odometry, which is obtained by optimizing errors of visual reprojection and IMU measurements, serves as an initial guess for lidar scan-matching and introduces constraints into the factor graph. 
After point cloud de-skewing using the IMU measurements, the LIS extracts lidar edge and planar features and matches them to a feature map that is maintained in a sliding window. The estimated system state in the LIS can be sent to the VIS to facilitate its initialization.
For loop closure, candidate matches are first identified by the VIS and further optimized by the LIS. The constraints from visual odometry, lidar odometry, IMU preintegration \cite{forster2016manifold}, and loop closure are optimized jointly in the factor graph. Lastly, the optimized IMU bias terms are leveraged to propagate IMU measurements for pose estimation at the IMU rate. The main contributions of our work can be summarized as follows:
\begin{itemize}
	\item A tightly-coupled LVIO framework built atop a factor graph, which achieves both multi-sensor fusion \textit{and} global optimization aided by place recognition.
	\item Our framework bypasses failed sub-systems via failure detection, making it robust to sensor degradation.
	\item Our framework is extensively validated with data gathered across varied scales, platforms, and environments.
\end{itemize}
Our work is novel from a systems standpoint, representing a unique integration of the state-of-the-art in VIO and LIO to achieve an LVIO system offering improved robustness and accuracy.
We hope that our system can serve as a solid baseline which others can easily build on to advance the state-of-the-art in lidar-visual-inertial odometry.

\section{Lidar Visual Inertial Odometry via Smoothing and Mapping}

\subsection{System Overview}

An overview of the proposed lidar-visual-inertial system, which receives inputs from a 3D lidar, a monocular camera, and an IMU, is shown in Figure \ref{fig::system-overview}. Our framework is composed of two key sub-systems: a visual-inertial system (VIS) and a lidar-inertial system (LIS). The VIS processes images and IMU measurements, with lidar measurements being optional. Visual odometry is obtained by minimizing the joint residuals of visual and IMU measurements. The LIS extracts lidar features and performs lidar odometry by matching the extracted features with a feature map. The feature map is maintained in a sliding-window manner for real-time performance. Lastly, the state estimation problem, which can be formulated as a maximum a posteriori (MAP) problem, is solved by jointly optimizing the contributions of IMU preintegration constraints, visual odometry constraints, lidar odometry constraints, and loop closure constraints in a factor graph using iSAM2 \cite{kaess2012isam2}. Note that the multi-sensor graph optimization employed in the LIS is intended to reduce data exchange and improve system efficiency. 

\subsection{Visual-Inertial System}
\label{sec::vio-system}

\begin{figure}[t!]
	\centering
	\includegraphics[width=.95\columnwidth]{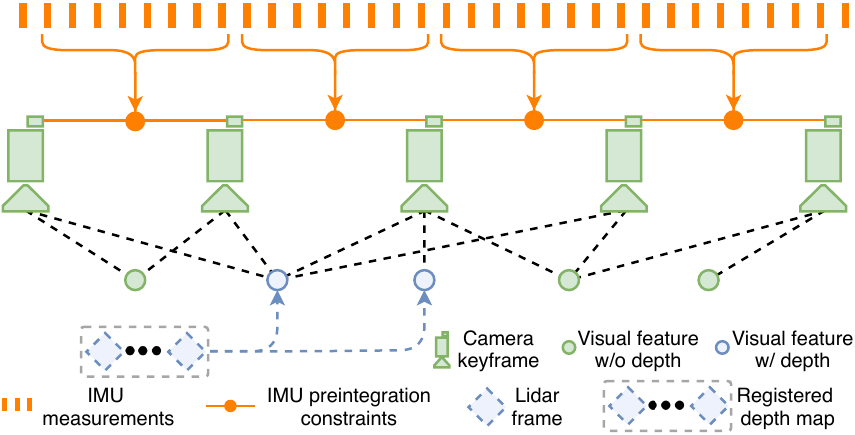}
	\caption{The framework of our visual-inertial system. The system optimizes residuals from IMU preintegration, visual measurements without depth, and visual measurements with depth.}
	\label{fig::vio-system}
	\vspace{-5mm}
\end{figure}

We adapt the processing pipeline from \cite{qin2018vins} for our VIS, which is shown in Figure \ref{fig::vio-system}. The visual features are detected using a corner detector \cite{shi1994good} and tracked by the Kanade–Lucas–Tomasi algorithm \cite{lucas1981iterative}. Upon VIS initialization, we register lidar frames using visual odometry and obtain a sparse depth image for feature depth estimation. The system performs bundle adjustment 
in a sliding-window setting, where system state $\bx \in \bX$ can be written as:
\begin{align}
	&\bx = {[\; \bR, \;\bp, \;\bv, \;\bb \;]}. \nonumber
\end{align}
$\bR \in SO(3)$ is the rotation matrix, $\bp \in \mathbb{R}^{3}$ is the position vector, $\bv$ is the speed, and $\bb = {[\;\bb_{a}, \bb_{w} \;]}$ is the IMU bias. $\bb_{a}$ and $\bb_{w}$ are the bias vectors for acceleration and angular velocity respectively. The transformation $\bT \in SE(3)$ from sensor body frame $\bold{B}$ to world frame $\bold{W}$ is represented as $\bT = [\bR | \bp]$. In the following sections, we give detailed procedures for improving VIS initialization and feature depth estimation. Due to space limitations, we refer readers to \cite{qin2018vins} for further details, such as the implementation of residuals.


\subsubsection{Initialization}
Optimization-based VIO usually suffers from divergence due to solving a highly-nonlinear problem at initialization. The quality of initialization heavily depends on two factors: the initial sensor movement and the accuracy of the IMU parameters. In practice, we find that \cite{qin2018vins} often fails to initialize when the sensor travels at a small or constant velocity. This is due to the fact that metric scale is not observable when acceleration excitation is not large enough. The IMU parameters include a slowly varying bias and white noise, which effect both the raw acceleration and angular velocity measurements. A good guess of these parameters at initialization helps the optimization converge faster.

To improve the robustness of our VIS initialization, we leverage the estimated system state $\bx$ and IMU bias $\bb$ from the LIS. Because depth is directly observable from a lidar, we first initialize the LIS and obtain $\bx$ and $\bb$. Then we interpolate and associate them to each image keyframe based on the image timestamp. Note that the IMU bias is assumed to be constant between two image keyframes. Finally, the estimated $\bx$ and $\bb$ from the LIS are used as the initial guess for VIS initialization, which significantly improves initialization speed and robustness. A comparison of VIS initialization with and without the help of the LIS can be found in the supplementary video\footnote{https://youtu.be/8CTl07D6Ibc}. 


\subsubsection{Feature depth association}

\begin{figure}[t!]
	\centering
	\subfigure[Depth association]{\includegraphics[width=.63\columnwidth]{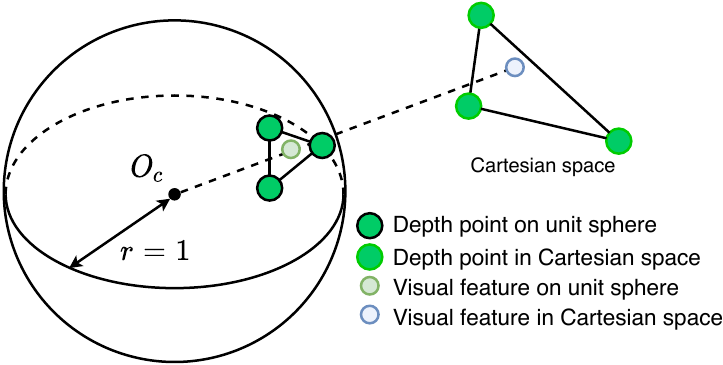}}
	\subfigure[Depth validation]{\includegraphics[width=.35\columnwidth]{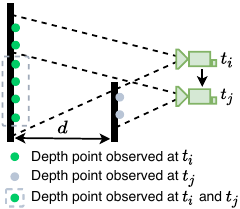}}
	\vspace{-2mm}
	\caption{Visual feature depth association.}
	\label{fig::vio-depth-association}
	\vspace{-0mm}
\end{figure}

Upon the initialization of the VIS, we register lidar frames to the camera frame using the estimated visual odometry. Due to the fact that a modern 3D lidar often yields sparse scans, we stack multiple lidar frames to obtain a dense depth map. To associate a feature with a depth value, we first project visual features and lidar depth points on a unit sphere that is centered at the camera. Depth points are then downsampled and stored using their polar coordinates for a constant density on the sphere. We find the nearest three depth points on the sphere for a visual feature by searching a two‐dimensional K‐D tree using the polar coordinates of the visual feature. At last, the feature depth is the length of the line formed by the visual feature and camera center $O_{c}$, which intersects the plane formed by the three depth points in Cartesian space. A visualization of this process can be found in Figure \ref{fig::vio-depth-association}(a), where the feature depth is the length of the dashed straight line.

We further validate the associated feature depth by checking the distance among the three nearest depth points. This is because stacking lidar frames from different timestamps may result in depth ambiguity from different objects. An illustration of such a case is shown in Figure \ref{fig::vio-depth-association}(b). The depth points observed at time $t_i$ are depicted in green. The camera is moved to a new position at $t_j$ and observes new depth points that are colored gray. However, the depth points at $t_i$, which are circled by a dashed gray line, may still be observable at $t_j$ due to lidar frame stacking. Associating feature depth using depth points from different objects results in inaccurate estimation. Similar to \cite{zhang2018laser}, we reject such estimations by checking the maximum distance among depth points for a feature. If the maximum distance is larger than 2m, the feature has no depth associated.

\begin{figure}[t!]
	\centering
    \subfigure[]{\includegraphics[width=.45\columnwidth]{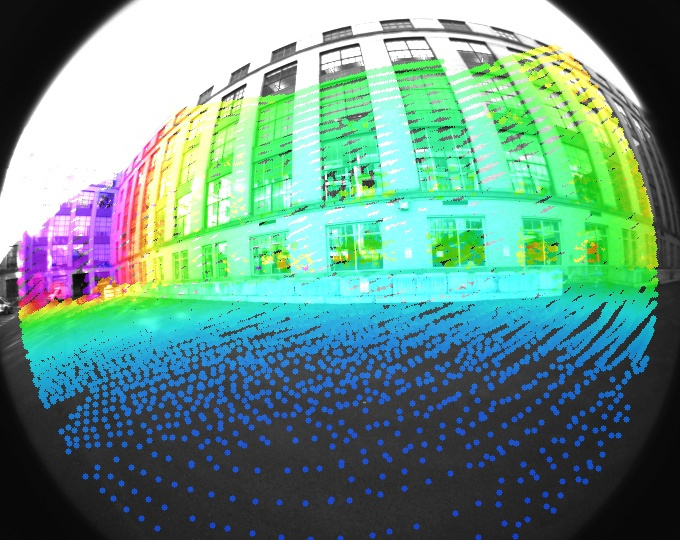}}
	\subfigure[]{\includegraphics[width=.45\columnwidth]{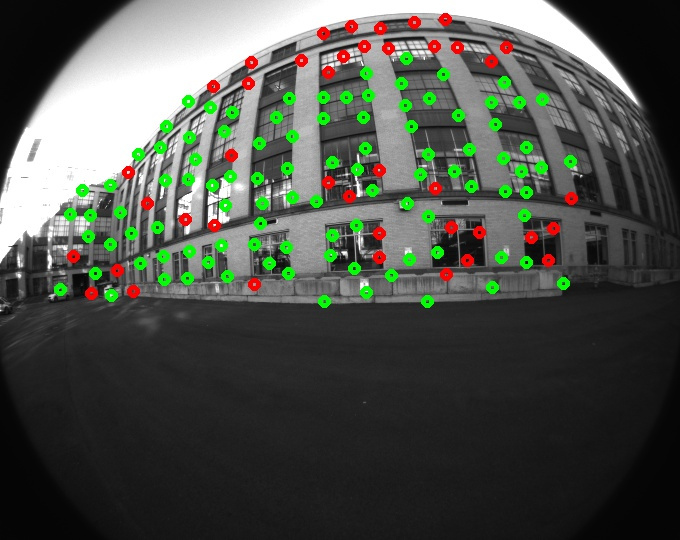}}
    \subfigure[]{\includegraphics[width=.45\columnwidth]{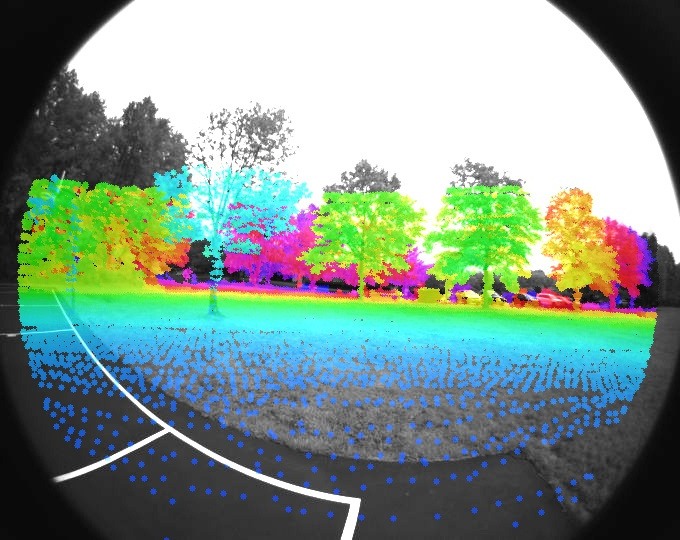}}
	\subfigure[]{\includegraphics[width=.45\columnwidth]{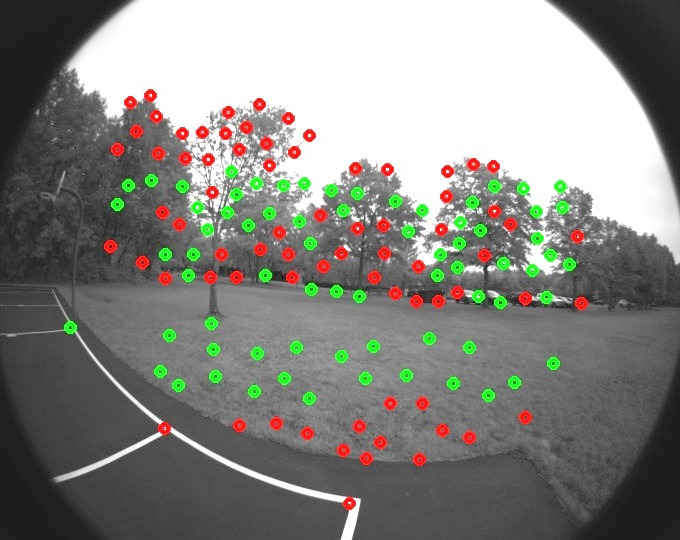}}
	\vspace{-2mm} \caption{Registered depth map and visual features. In (a) and (c), the color variation of the depth map indicates depth change. In (b) and (d), green points are visual features that are successfully associated with depth. Features that fail the depth association process are red.}
	\label{fig::vio-depth-demo}
	\vspace{-5mm}
\end{figure}

A demonstration of the registered depth map and visual features is shown in Figure \ref{fig::vio-depth-demo}. In Figure \ref{fig::vio-depth-demo} (a) and (c), the depth points that are registered using visual odometry are projected onto the camera images. In Figure \ref{fig::vio-depth-demo} (b) and (d), the visual features that are successfully associated with depth are colored green. Note that though the depth map covers the majority of the image in Figure \ref{fig::vio-depth-demo}(a), many features in \ref{fig::vio-depth-demo}(b) that are located on the corners of windows lack 
depth association due to validation check failure.

\subsubsection{Failure detection}

The VIS suffers from failure due to aggressive motion, illumination change, and texture-less environments. When a robot undergoes aggressive motion or enters a texture-less environment, the number of tracked features decreases greatly. Insufficient features may lead to optimization divergence. We also notice that a large IMU bias is estimated when the VIS fails. Thus we report VIS failure when the number of tracked features is below a threshold, or when the estimated IMU bias exceeds a threshold. Active failure detection is necessary for our system so that its failure does not corrupt the function of the LIS. Once a failure is detected, the VIS re-initializes and informs the LIS. 

\subsubsection{Loop closure detection}

We utilize DBoW2 \cite{galvez2012bags} for loop closure detection. For each new image keyframe, we extract BRIEF descriptors \cite{calonder2010brief} and match them with previously extracted descriptors. The image timestamps of loop closure candidates returned by DBoW2 are sent to the LIS for further validation.

\subsection{Lidar-Inertial System}
\label{sec::lio-system}

\begin{figure}[t!]
	\centering
	\includegraphics[width=.99\columnwidth]{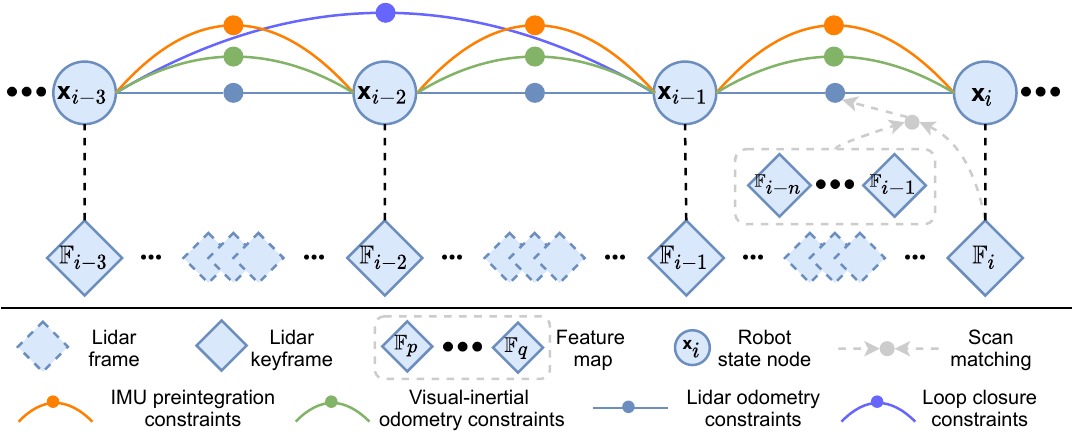}
	\caption{The framework of our lidar-inertial system. The system maintains a factor graph that has four types of constraints.}
	\label{fig::lio-system}
	\vspace{-0mm}
\end{figure}

As is shown in Figure \ref{fig::lio-system}, the proposed lidar-inertial system, which is adapted from \cite{shan2020lio}, maintains a factor graph for global pose optimization. Four types of constraints, IMU preintegration constraints, visual odometry constraints, lidar odometry constraints, and loop closure constraints, are added to the graph and optimized jointly. 
Lidar odometry constraints are derived from scan-matching, where we match the current lidar keyframe to a global feature map. The candidates for loop closure constraints are first provided by the VIS and then further optimized by scan-matching. We maintain a sliding window of lidar keyframes for the feature map, which guarantees bounded computational complexity. A new lidar keyframe is selected when the change of the robot pose exceeds a threshold. The intermittent lidar frames lying between pairs of keyframes are discarded. Upon the selection of a new lidar keyframe, a new robot state $\bx$ is added to the factor graph as a node. Adding the keyframes in this way not only achieves a balance between memory consumption and map density, but also helps maintain a relatively sparse factor graph for real-time optimization. Due to space limitations, we refer the readers to \cite{shan2020lio} for implementation details. In the following sections, we focus on new procedures that improve system robustness.

\subsubsection{Initial guess}

We find the initial guess plays a critical role in the success of scan-matching, especially when the sensor undergoes aggressive motions. The source of initial guess is different before and after LIS initialization.

Before LIS initialization, we assume the robot starts from a static position with zero velocity. Then we integrate raw IMU measurements assuming the bias and noise are zero-valued. The integrated translational and rotational change between two lidar keyframes produce the initial guess for scan-matching. We find this approach can successfully initialize the system in challenging conditions when the initial linear velocity is smaller than 10 $m/s$ and the angular velocity is smaller than 180 $^\circ/s$. Once the LIS is initialized, we estimate the IMU bias, robot pose, and velocity in the factor graph. Then we send them to the VIS to aid its initialization.

After LIS initialization, we can obtain initial guesses from two sources: integrated IMU measurements with corrected bias, and the VIS. We use visual-inertial odometry as an initial guess when it is available. If the VIS reports failure, we then switch to IMU measurements for the initial guess. These procedures increase the initial guess accuracy and robustness in both texture-rich and texture-less environments.

\subsubsection{Failure detection}

\begin{figure}[t!]
	\centering
	\subfigure[]{\includegraphics[width=.45\columnwidth]{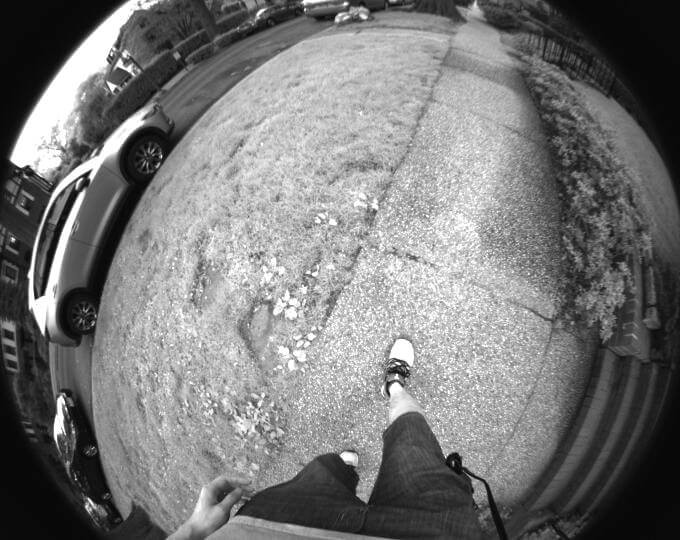}}
	\subfigure[]{\includegraphics[width=.45\columnwidth]{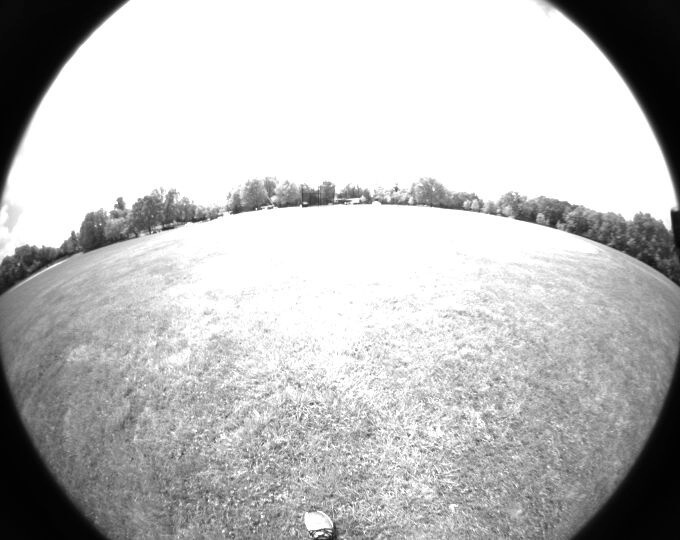}}
	\subfigure[]{\includegraphics[width=.45\columnwidth]{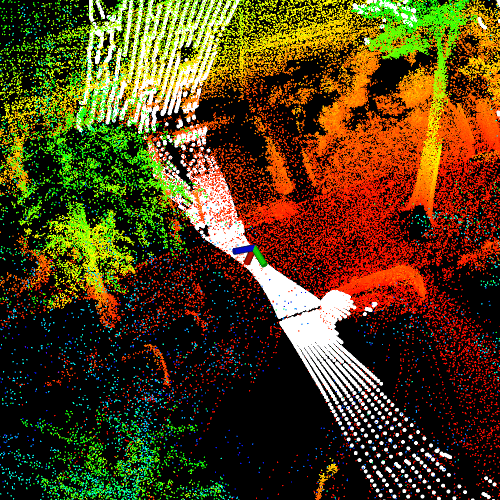}}
	\subfigure[]{\includegraphics[width=.45\columnwidth]{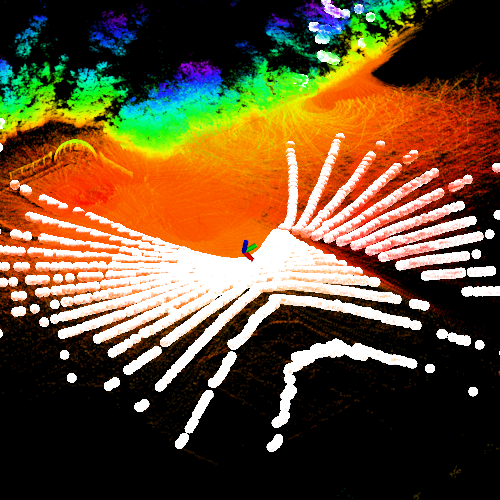}}
	\vspace{-0mm}
	\caption{Degenerate mapping scenarios where scan-matching is ill-constrained. In (a) and (c), the lidar is placed facing the ground. In (b) and (d), the lidar is in a flat, open, and structure-less environment. White points indicate the contents of the lidar scan. Color variation indicates elevation change.}
	\label{fig::lio-degeneration}
	\vspace{-0mm}
\end{figure}

Though lidar can capture the fine details of an environment at long ranges, it still encounters degraded scenarios where scan-matching is ill-constrained. Such scenarios are depicted in Figure \ref{fig::lio-degeneration}. We adapt the method from \cite{zhang2016degeneracy} for LIS failure detection. The non-linear optimization problem in scan-matching can be formulated as solving a linear problem iteratively:
\begin{align}
	\min_{\bT}
	    \parallel   \bold{A} \bT - \mathbf{b}     \parallel ^{2},
\end{align}
where $\bold{A}$ and $\mathbf{b}$ are obtained from linearization at $\bT$. The LIS reports failure when the smallest eigenvalue of $\bold{A}^{\mathsf{T}}\bold{A}$ is smaller than a threshold at the first iteration of optimization. Lidar odometry constraints are not added to the factor graph when failure happens. We refer the reader to \cite{zhang2016degeneracy} for the detailed analysis upon which these assumptions are based. 


\section{Experiments}
\label{sec::experiment}

We now describe a series of experiments to validate the proposed framework on three self-gathered datasets, which are referred to as $Urban$, $Jackal$, and $Handheld$. The details of these datasets are provided in the following sections. Our sensor suite for data-gathering includes a Velodyne VLP-16 lidar, a FLIR BFS-U3-04S2M-CS camera, a MicroStrain 3DM-GX5-25 IMU, and a Reach RS+ GPS (for ground truth). We compare the proposed framework with open-sourced solutions, which include VINS-Mono, LOAM, LIO-mapping, LINS, and LIO-SAM. All the methods are implemented in C++ and executed on a laptop with an Intel i7-10710U in Ubuntu Linux. Our implementation of LVI-SAM and datasets are available at the link below\footnote{\url{https://git.io/lvi-sam}}.

\subsection{Ablation Study}

\begin{figure}[t]
	\centering
	\includegraphics[width=.95\columnwidth]{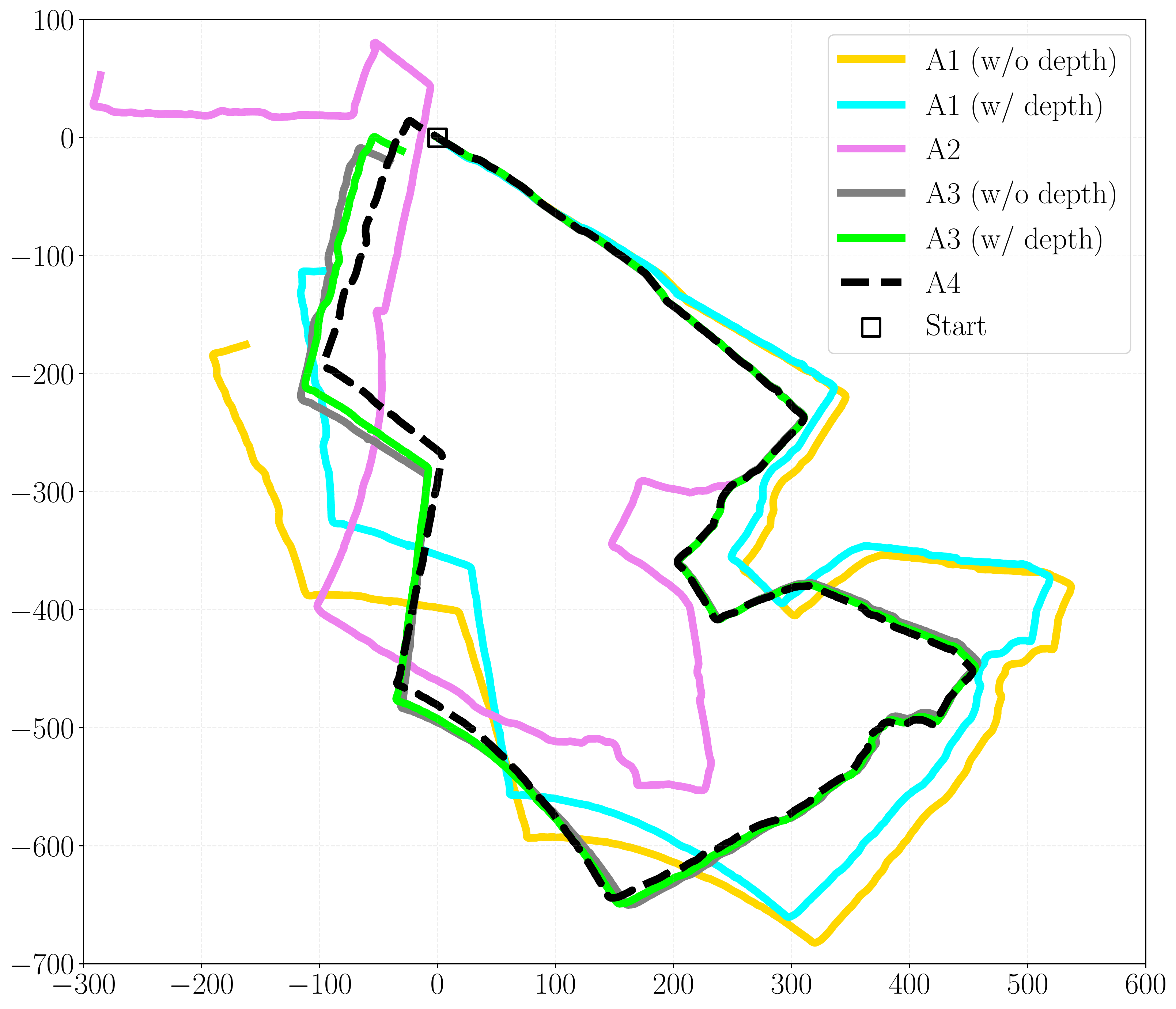}
	\caption{Trajectories of ablation study using Urban dataset.}
	\label{fig::ablation-path}
	\vspace{-0mm}
\end{figure}
We show how the design of each module in the system affects the performance of the proposed framework using the $Urban$ dataset. This dataset, which features buildings, parked and moving cars, pedestrians, cyclists, and vegetation, is gathered by the operator walking and carrying the sensor suite. We also deliberately place the sensor suite in challenging positions (Figure \ref{fig::lio-degeneration}(a)) to validate the robustness of the system in degraded scenarios. Due to dense overhead vegetation, the region is GPS-denied. We start and end the data-gathering process at the same position to validate end-to-end translation and rotation errors, provided in Table \ref{tab::benchmark-ablation}.

\begin{table}[t]
\centering
	\caption{End-to-end translation and rotation errors.}
	\label{tab::benchmark-ablation}
	\resizebox{0.99\columnwidth}{!}{
    \begin{tabular}{ccccccc}
    \toprule
    Error type & \begin{tabular}[c]{@{}c@{}}A1\\ (w/o depth)\end{tabular} & \begin{tabular}[c]{@{}c@{}}A1\\ (w/ depth)\end{tabular} & A2 & \begin{tabular}[c]{@{}c@{}}A3\\ (w/o depth)\end{tabular} & \begin{tabular}[c]{@{}c@{}}A3\\ (w/ depth)\end{tabular} & A4 \\
    \midrule
    Translation (m) & 239.19 & 142.12 & 290.43 & 45.42 & 32.18 & 0.28 \\
    Rotation (degree) & 60.72 & 39.52 & 116.50 & 6.41 & 7.66 & 5.77 \\
    \bottomrule
    \end{tabular}}
    \vspace{-0mm}
\end{table}
\subsubsection{A1 - Effect of including feature depth information from lidar for visual-inertial odometry}
We disable the scan-matching in the LIS and perform pose estimation solely relying on the VIS. The resulting trajectories with and without enabling depth registration are labeled as A1 in Figure \ref{fig::ablation-path}. The trajectory direction is clock-wise. The end-to-end pose errors, shown in Table \ref{tab::benchmark-ablation}, are greatly reduced when associating depth with visual features. 

\subsubsection{A2 - Effect of including visual-inertial odometry}
We disable the VIS and use only the LIS for pose estimation. The trajectory, which is labeled A2 in Figure \ref{fig::ablation-path}, diverges a couple of times when a degraded scenario is encountered.

\subsubsection{A3 - Effect of including feature depth information from lidar for lidar-visual-inertial odometry}
We now use the VIS and LIS together, toggling the depth registration module in the VIS to compare the resulting LVIO trajectories. With the help of depth for visual features, the translation error is further reduced by 29$\%$, from 45.42 m to 32.18 m. Note that the loop closure detection is disabled in this test to validate the system in a pure odometry mode.

\subsubsection{A4 - Effect of including visual loop closure detection}
We eliminate the drift of the system by enabling the loop closure detection function in the VIS. The final trajectory when every module is enabled in the framework is labeled as A4 in Figure \ref{fig::ablation-path}.

\begin{figure}[t!]
	\centering
	\subfigure[Jackal dataset environment]{\includegraphics[width=.445\columnwidth]{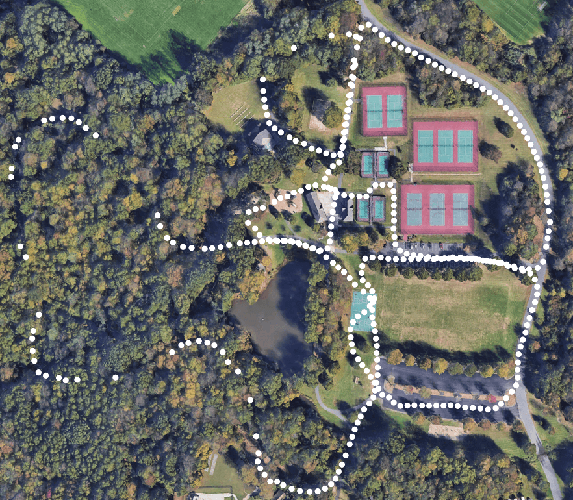}}
	\subfigure[Handheld dataset environment]{\includegraphics[width=.51\columnwidth]{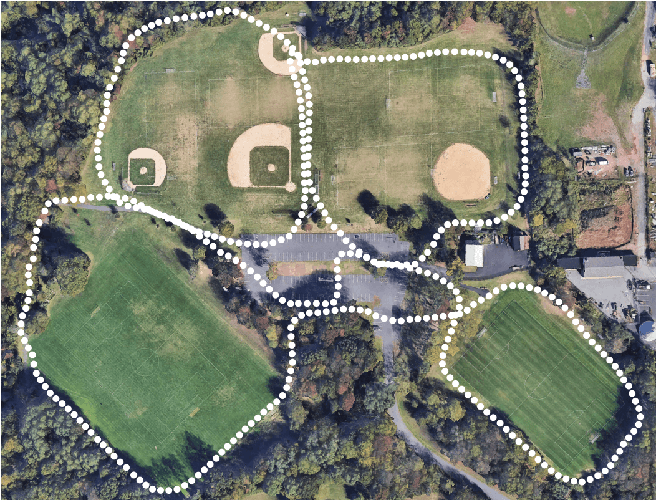}}
	\vspace{-0mm}
	\caption{Satellite images where the Jackal and Handheld datasets were gathered. White dots indicate GPS availability in the datasets.}
	\label{fig::dataset-earth}
	\vspace{-3mm}
\end{figure}

\begin{figure*}[t!]
	\centering
	\subfigure[Trajectories of Jackal dataset]{\includegraphics[width=.99\textwidth]{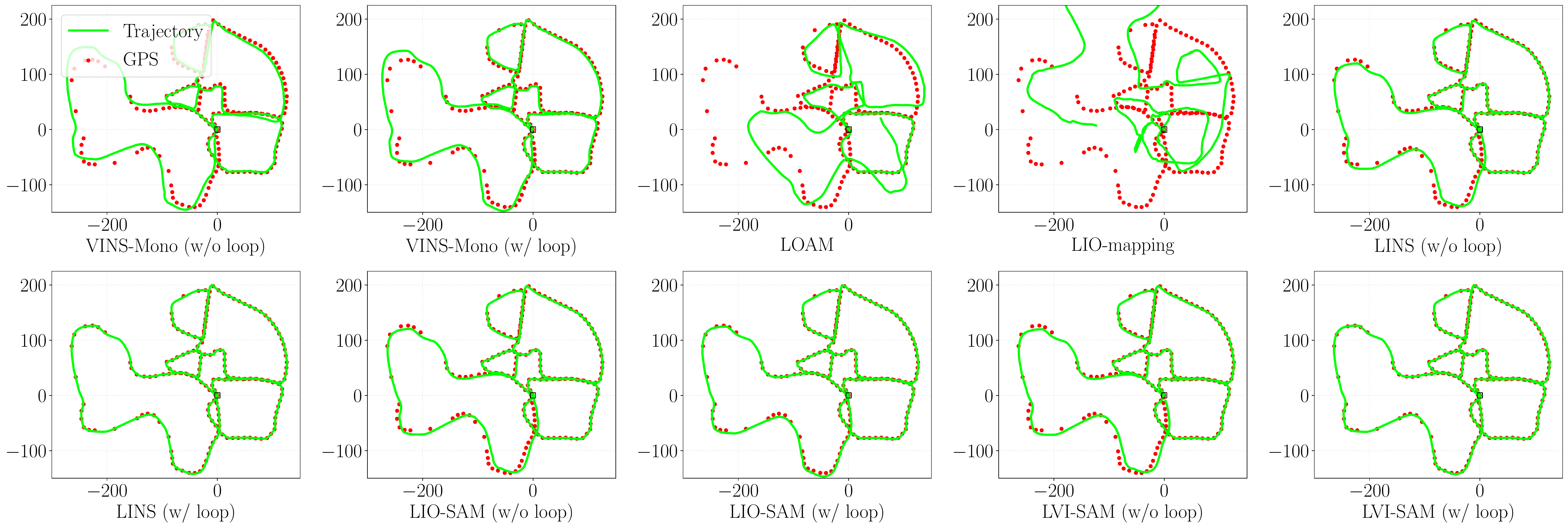}}
	\subfigure[Trajectories of Handheld dataset]{\includegraphics[width=.99\textwidth]{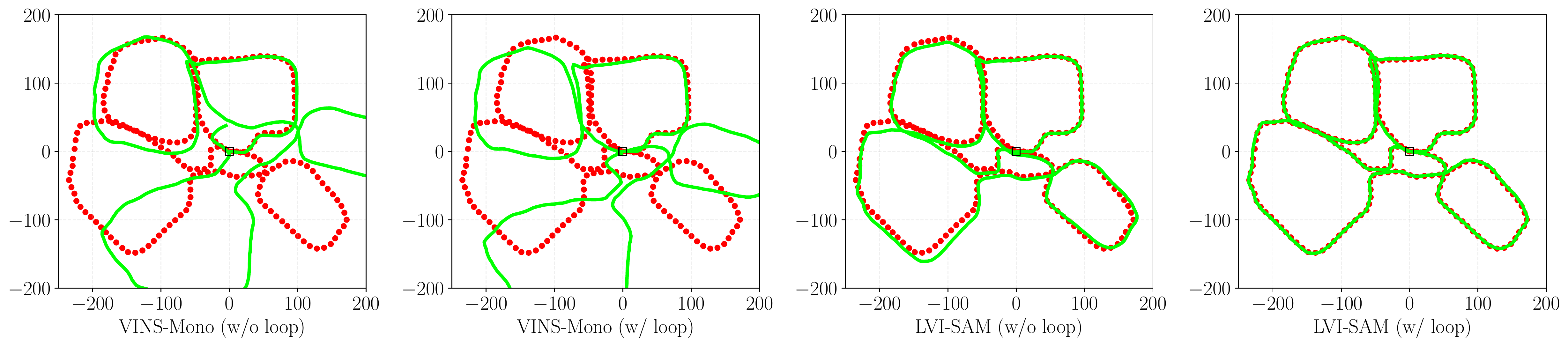}}
	\vspace{-0mm}
	\caption{Trajectories of various methods using the Jackal and Handheld datasets. The green line depicts the resulting trajectory from each compared method. The GPS positioning measurements, which serve as the ground truth, are shown with red dots. Trajectories of LOAM, LIO-mapping, LINS, and LIO-SAM are not shown due to their failure to generate meaningful results.}
	\label{fig::dataset-path}
	\vspace{-0mm}
\end{figure*}

\begin{table*}[t]
	\centering
	\caption{Quantitative comparison of Jackal and Handheld datasets using various methods.}
	\label{tab::benchmark-error}
	\resizebox{0.99\textwidth}{!}{
	\begin{tabular}{cccccccccccc}
		\toprule
		Dataset & Error type & \begin{tabular}[c]{@{}c@{}}VINS\\ (w/o loop)\end{tabular} & \begin{tabular}[c]{@{}c@{}}VINS\\ (w/ loop)\end{tabular} & LOAM & LIO-mapping & \begin{tabular}[c]{@{}c@{}}LINS\\ (w/o loop)\end{tabular} & \begin{tabular}[c]{@{}c@{}}LINS\\ (w/ loop)\end{tabular} & \begin{tabular}[c]{@{}c@{}}LIO-SAM\\ (w/o loop)\end{tabular} & \begin{tabular}[c]{@{}c@{}}LIO-SAM\\ (w/ loop)\end{tabular} & \begin{tabular}[c]{@{}c@{}}LVI-SAM\\ (w/o loop)\end{tabular} & \begin{tabular}[c]{@{}l@{}}LVI-SAM\\ (w/ loop)\end{tabular} \\
		\midrule[1pt]
		\multirow{3}{*}{Jackal} & \begin{tabular}[c]{@{}c@{}}RMSE w.r.t GPS (m)\end{tabular} & 8.58 & 4.49 & 44.92 & 127.05 & 3.95 & 0.77 & 3.54 & 1.52 & 4.05 & \textbf{0.67} \\
		\cmidrule{2-12}
		& \begin{tabular}[c]{@{}c@{}}Translation (m)\end{tabular} & 10.82 & 11.86 & 61.73 & 123.22 & 7.37 & \textbf{0.09} & 5.48 & 0.12 & 4.69 & 0.11 \\
		\cmidrule{2-12}
		& \begin{tabular}[c]{@{}c@{}}Rotation (degree)\end{tabular} & 9.98 & 12.79 & 61.41 & 139.23 & 4.80 & 2.07 & 2.18 & 2.64 & 2.28 & \textbf{1.52} \\
		\midrule[1pt]
		\multirow{3}{*}{Handheld} & \begin{tabular}[c]{@{}c@{}}RMSE w.r.t GPS (m)\end{tabular} & 87.53 & 73.07 & Fail & Fail & Fail & Fail & 53.62 & Fail & 7.87 & \textbf{0.83} \\
		\cmidrule{2-12}
		& \begin{tabular}[c]{@{}c@{}}Translation (m)\end{tabular} & 40.65 & 1.87 & Fail & Fail & Fail & Fail & 58.91 & Fail & 7.57 & \textbf{0.27} \\
		\cmidrule{2-12}
		& \begin{tabular}[c]{@{}c@{}}Rotation (degree)\end{tabular} & 53.48 & 52.09 & Fail & Fail & Fail & Fail & 20.17 & Fail & 24.82 & \textbf{3.82} \\
		\bottomrule
	\end{tabular}}
	\vspace{-0mm}
\end{table*}

\subsection{Jackal Dataset}

The $Jackal$ dataset is gathered by mounting the sensor suite on a Clearpath Jackal unmanned ground vehicle (UGV). We manually drive the robot in a feature-rich environment, beginning and ending at the same position. The environment, shown in Figure \ref{fig::dataset-earth}(a), features structures, vegetation, and various road surfaces. The regions where GPS reception is available are marked with white dots.

We compare various methods and show their trajectories in Figure \ref{fig::dataset-path}(a). We further validate the accuracy of methods that have loop closure functionality by manually disabling and enabling it. The benchmarking results are shown in Table \ref{tab::benchmark-error}. LVI-SAM achieves the lowest average root mean square error (RMSE) with respect to the GPS measurements, which are treated as ground truth. The lowest end-to-end translation error is achieved by LINS, which is adapted from LeGO-LOAM\cite{shan2018lego} and specifically designed for UGV operations. The lowest end-to-end rotational error is again achieved by LVI-SAM.

\subsection{Handheld Dataset}

The $Handheld$ dataset is gathered by an operator carrying the sensor suite walking around several open fields, which are shown in Figure \ref{fig::dataset-earth}(b). The dataset also starts and ends at the same position. We increase the challenge of this dataset by passing an open baseball field, which is located at top center of the image. When passing this field, the main observations gathered by the camera and lidar feature grass and a ground plane, respectively (Fig. \ref{fig::lio-degeneration} (b) and (d)). Due to the aforementioned degeneracy problem, all the lidar-based methods fail to generate meaningful results. The proposed framework, LVI-SAM, successfully finishes the test with or without loop closures enabled, achieving the lowest errors over all three benchmarking criteria featured in Table \ref{tab::benchmark-error}.

\section{Conclusions}

We have proposed LVI-SAM, a framework for tightly-coupled lidar-visual-inertial odometry via smoothing and mapping, for performing real-time state estimation and mapping in complex environments. The proposed framework is composed of two sub-systems: a visual-inertial system and a lidar-inertial system. The two sub-systems are designed to interact in a tightly-coupled manner to improve system robustness and accuracy. Through evaluations on datasets over various scales, platforms, and environments, our system shows comparable or better accuracy than the existing publicly available methods. We hope that our system will serve as a solid baseline which others can easily build on to advance the state of the art in lidar-visual-inertial odometry.

\section*{Acknowledgement}

This work was supported by Amsterdam Institute for Advanced Metropolitan Solutions, Amsterdam, the Netherlands.

\bibliographystyle{IEEEtran}
\bibliography{ICRA_2021_SAM}
\end{document}